\documentclass[preprint,review,10pt]{elsarticle}

\usepackage[T1]{fontenc}
\usepackage{graphicx}
\usepackage{booktabs}
\usepackage{multirow}
\usepackage{adjustbox}
\usepackage{subcaption}
\usepackage{placeins}
\usepackage{float}
\usepackage{grffile}
\usepackage{caption}
\usepackage{amsmath,amssymb}
\usepackage{xspace}
\usepackage[ruled,vlined,linesnumbered]{algorithm2e}

\SetAlgoLined
\SetAlgoVlined
\SetAlgoSkip{4pt}
\SetAlgoInsideSkip{1.5pt}
\SetKwComment{Comment}{\(\triangleright\)~}{}
\SetCommentSty{textnormal}

\makeatletter
\def\fps@figure{!htbp}
\def\fps@table{!htbp}
\makeatother

\begin{document}

\begin{frontmatter}

\title{AIDA-ReID: Adaptive Intermediate Domain Adaptation for Generalizable and Source-Free Person Re-Identification}

\author[1]{Sundas Iqbal}
\ead{sundasiqbal058@gmail.com}

\author[1,2]{Qing Tian\corref{cor1}}
\ead{tianqing@nuist.edu.cn}

\author[3]{Danish Ali}
\ead{danishalikhan545@gmail.com}

\author[4]{Jianping Gou}
\ead{cherish.gjp@gmail.com}

\author[5]{Weihua Ou}
\ead{ouwei-hua@gznu.edu.cn}

\address[1]{School of Software, Nanjing University of Information Science and Technology, Nanjing 210044, China}
\address[2]{Wuxi Institute of Technology, Nanjing University of Information Science and Technology, Wuxi 214000, China}
\address[3]{School of Computer Science, Wuhan University, Wuhan 430070, China}
\address[4]{School of Computer and Information Science, Southwest University, Chongqing 400715, China}
\address[5]{School of Big Data and Computer Science, Guizhou Normal University, Guiyang 550025, China}

\cortext[cor1]{Corresponding author: Qing Tian (tianqing@nuist.edu.cn)}

\begin{abstract}
Person re-identification (Re-ID) aims to match images of the same individual across non-overlapping camera views and remains challenging due to domain shifts caused by variations in illumination, background, camera characteristics and population distributions. Although supervised models perform well under matched training and testing conditions, their performance degrades significantly when deployed in unseen environments. Existing intermediate-domain approaches such as IDM and IDM++ alleviate this gap by constructing bridge feature distributions between domains; however, they rely on fixed mixing strategies and joint source--target access limiting their applicability to multi-source and source-free settings. To address these limitations, this paper proposes Adaptive Intermediate Domain Adaptation (AIDA) also referred to as Source-Free Multi-Source Intermediate Domain Adaptation (SF-MIDA). The proposed framework treats intermediate-domain learning as a dynamically regulated process, where feature mixing and regularization strength are adaptively controlled using feedback signals derived from model uncertainty and training stability. A multi-source intermediate domain generator synthesizes diverse intermediate representations, while a pseudo-mirror regularization strategy preserves identity consistency under domain perturbations. Extensive experiments across domain generalization and source-free settings demonstrate the effectiveness of the proposed framework.
\end{abstract}


\end{frontmatter}

\section{Introduction}

Person re-identification (Re-ID) aims to match images of the same individual captured by non-overlapping cameras across different viewpoints and environments \cite{ahmed2026maformer}. It is a core component of intelligent video surveillance, forensic analysis and cross-camera tracking systems \cite{tian2025cross}. Although supervised deep learning models achieve strong performance on benchmark datasets such as Market-1501, DukeMTMC-ReID and MSMT17, their effectiveness degrades noticeably when deployed in unseen environments \cite{zhou2025dynamicmix}. This performance drop is primarily caused by domain shift, where variations in illumination, background, resolution, camera characteristics and population distributions induce substantial discrepancies between training and deployment data \cite{zhang2025cross}. To address domain shift, two major paradigms have been widely explored: unsupervised domain adaptation (UDA) and domain generalization (DG). UDA methods reduce distribution gaps by leveraging labeled source data together with unlabeled target data through clustering, adversarial learning or pseudo-label refinement \cite{ran2025camera}. However, their reliance on target-domain data limits scalability and raises practical and privacy concerns in real-world deployment\cite{chen2023unsupervised}. DG instead focuses on learning domain-invariant representations from one or multiple labeled source domains without access to target data. While DG offers greater deployment flexibility existing approaches often struggle to model the continuous evolution of feature distributions across heterogeneous domains leading to limited generalization under severe domain shifts. 

From a geometric perspective, source and target domains can be viewed as distant regions on a high-dimensional manifold \cite{tian2025unsupervised}. Direct alignment between these regions often induces abrupt transitions and unstable optimization. To mitigate this issue, intermediate-domain learning constructs bridge distributions that gradually connect source and target domains \cite{zhang2025sparse}. Representative approaches such as the Intermediate Domain Module (IDM) \cite{dai2021idm} and its extension IDM++ \cite{dai2025idmpp} synthesize intermediate feature representations via feature-statistics mixing with IDM++ further incorporating mirror generation based on Adaptive Instance Normalization (AdaIN) to preserve identity semantics \cite{du2026msc}. Despite their effectiveness, existing intermediate-domain methods suffer from three key limitations: (1) they require simultaneous access to source and target data which precludes source-free deployment; (2) they are primarily designed for single source--target adaptation and cannot fully exploit heterogeneous multi-source information and (3) they rely on fixed interpolation strategies and static loss weights preventing adaptive regulation under evolving training dynamics. These observations indicate that intermediate-domain learning for person Re-ID should be treated as a dynamic process rather than a fixed design choice particularly in multi-source and source-free settings where corrective adaptation at deployment time is infeasible.

Motivated by this insight, we propose AIDA/SF-MIDA, a feedback-regulated framework for generalizable person Re-ID. The proposed approach addresses three fundamental challenges in multi-source and source-free Re-ID constructing adaptive intermediate domains without relying on target data preserving identity semantics during continuous domain transitions and dynamically regulating domain interpolation and training stability using internal feedback signals. To this end, AIDA/SF-MIDA integrates three tightly coupled components. A Multi-Source Intermediate Domain Generator (MS-IDG) synthesizes intermediate representations across multiple source domains to expand the effective training distribution. A Pseudo-Mirror Regularization (PMR) mechanism enforces identity consistency under controlled domain perturbations. Complementing these components, a Dynamic Feedback Controller (DFC) adaptively adjusts interpolation strength and regularization weights based on prediction uncertainty and optimization stability. Through feedback-driven regulation AIDA/SF-MIDA enables smooth and stable domain transitions while maintaining discriminative identity representations without accessing target-domain data during deployment. The main contributions of this work are summarized as follows:
\begin{itemize}
    \item We propose AIDA/SF-MIDA, a unified feedback-regulated intermediate-domain learning framework that reformulates intermediate-domain adaptation as a dynamic control problem rather than a static feature-mixing process enabling robust multi-source generalization and source-free deployment for person re-identification.
    
    \item We introduce an MS-IDG that constructs a continuous family of adaptive intermediate domains by jointly modeling and interpolating feature statistics across heterogeneous source distributions, thereby expanding the effective training support beyond individual source domains and mitigating overfitting to domain-specific characteristics.
    
    \item We design a PMR mechanism that enforces both identity-level and relational consistency under intermediate-domain perturbations preserving embedding geometry and preventing semantic drift during aggressive domain interpolation.
    
    \item We develop a DFC that closes the learning loop by regulating intermediate-domain interpolation strength and regularization weights based on prediction uncertainty and optimization stability enabling self-regulated training that balances domain exploration and identity preservation under severe and unseen domain shifts.
\end{itemize}

The remainder of this paper is organized as follows.
Section~\ref{sec:related} reviews related work. Section~\ref{sec:method} presents the proposed AIDA/SF-MIDA framework. Section~\ref{sec:experiments} describes the experimental setup and evaluation protocols. Section~\ref{sec:results} reports the experimental results and analyses. Section~\ref{sec:discussion} discusses key observations and insights. Section~\ref{sec:conclusion} concludes the paper and outlines future directions.

\section{Related Work}
\label{sec:related}

Person Re-ID across disjoint camera networks remains challenging due to severe domain shifts caused by variations in illumination, viewpoint and background conditions. Existing research mainly addresses these issues through unsupervised domain adaptation, domain generalization, intermediate-domain learning and feedback-based regularization strategies.

\subsection{Unsupervised Domain Adaptation}

Unsupervised Domain Adaptation (UDA) aims to transfer discriminative knowledge from a labeled source domain to an unlabeled target domain. Early approaches such as ECN~\cite{zhong2019ecn} employed memory-based exemplar learning for cross-domain feature alignment, while subsequent studies explored adversarial learning and style-transfer strategies to mitigate domain discrepancies. Although these techniques improved adaptation performance, aggressive distribution alignment often introduced identity distortion or appearance artifacts.

Recent methods focus on pseudo-label refinement and self-training to improve target identity consistency. Representative approaches include HDNet~\cite{hao2022hdnet}, SGCL~\cite{wu2024sgcl} and COKD~\cite{wang2025cokd}, which utilize disentanglement, contrastive learning and knowledge distillation to enhance cross-domain representation learning. Other methods such as MLA~\cite{zheng2024mla} and IDENet~\cite{yang2025idenet} incorporate multi-level attention and equilibrium learning mechanisms to balance inter-domain generalization with intra-domain discrimination. Despite these advances, most UDA frameworks still rely on access to source data and assume a fixed source--target setting, limiting their applicability under domain generalization or source-free scenarios.

\subsection{Domain Generalization}

Domain Generalization (DG) aims to learn domain-invariant representations from one or multiple labeled source domains that generalize to unseen targets without accessing target data. Recent approaches explore meta-learning, style uncertainty modeling and episodic training strategies to simulate domain shifts during optimization and improve representation robustness. Multi-source frameworks further exploit complementary information across domains, while contrastive regularization enhances feature discrimination and reduces inter-domain variance. These strategies highlight the importance of jointly modeling domain diversity and representation consistency for reliable performance under unseen domain shifts.

Several studies further investigate the balance between discriminability and generalizability. Hu et al.~\cite{hu2024benchmark} introduced a comprehensive DG benchmark for person Re-ID, while Zhou et al.~\cite{zhou2024rethinkingdg} proposed adaptive normalization mechanisms to mitigate overfitting to source-specific styles. Wang et al.~\cite{wang2025lvitnet} developed LViT-Net, a lightweight vision transformer with feedback mechanisms that improves stability under domain shift and Niu et al.~\cite{niu2026fdgreid} explored federated DG through FDGReID to address domain drift and data privacy. Despite these advances, many DG methods still assume discrete domain boundaries making it difficult to capture gradual inter-domain transitions encountered in real-world scenarios.

\subsection{Intermediate-Domain, Regularization and Feedback-Based Learning}

Intermediate-domain and bridge-based learning aim to mitigate domain shift by constructing transitional feature distributions between source and target domains. Dai et al.~\cite{dai2021idm} introduced the Intermediate Domain Module (IDM) to generate intermediate feature representations, while its extension IDM++~\cite{dai2025idmpp} incorporated mirror generation with Adaptive Instance Normalization (AdaIN) to preserve identity semantics during domain interpolation. Related augmentation strategies such as StyleMix~\cite{kim2023stylemix} and MixStyle~\cite{zhou2024mixstyle} interpolate feature statistics to simulate virtual domains and improve generalization. Other approaches including AdaBDG~\cite{wu2024adabdg} and gradual alignment strategies~\cite{li2025gradualalign} further explore adaptive bridge-domain generation and scalable interpolation mechanisms in multi-source scenarios.

Regularization and feedback-based learning further enhance robustness under distribution shifts by constraining representation geometry and adapting training dynamics. Existing methods employ entropy minimization, feature consistency and contrastive regularization to mitigate cross-domain uncertainty. Representative examples include entropy-based feedback controllers~\cite{zhao2024adaptivefb}, feedback-driven transformer models~\cite{wang2025lvit} gradient-based regulation mechanisms~\cite{chen2025gradcontrol} and self-regularized meta-learning frameworks~\cite{lee2024selfreg}.

Despite these advances, most UDA and DG methods still rely on static interpolation or fixed pseudo-label assumptions and often require access to source or target data. In contrast, AIDA/SF-MIDA models intermediate-domain construction as a dynamic process that integrates multi-source domain mixing identity-preserving transitions and feedback-driven regulation to adapt to evolving training dynamics.

\section{Methodology} 
\label{sec:method}

\subsection{Problem Definition}
Person Re-ID aims to retrieve images of the same identity across disjoint camera views. In real-world deployments, Re-ID models often suffer from performance degradation due to domain shifts caused by variations in illumination, background, camera characteristics and population distributions. In this work, we study generalizable and source-free person Re-ID, where models are required to generalize to unseen environments without accessing target-domain data during training.

Let $\{\mathcal{D}_s^k\}_{k=1}^{K}$ denote $K$ labeled source domains. As defined in Eq.~(\ref{eq:source_domain}), the $k$-th source domain is
\begin{equation}
\mathcal{D}_s^k = \{(x_i^k, y_i^k)\}_{i=1}^{N_k}
\label{eq:source_domain}
\end{equation}
where $x_i^k\!\in\!\mathbb{R}^{H\times W\times 3}$ denotes a pedestrian image and $y_i^k\!\in\!\{1,\ldots,C_k\}$ its identity label. Identity label spaces across source domains are assumed to be disjoint, i.e., $C_k \cap C_{k'} = \emptyset$ for $k \neq k'$.

At test time, the trained model is deployed to an unseen target domain
$\mathcal{D}_u = \{x_j^u\}_{j=1}^{N_u}$, whose data distribution differs from all source domains and whose identity labels are unavailable. Under the source-free setting, source-domain data are inaccessible during adaptation, while only pretrained model parameters are retained; unlabeled target samples may optionally be used for target-only refinement. We adopt a standard embedding-based Re-ID paradigm, where an input image $x$ is mapped by a feature extractor $f_{\theta}(\cdot)$ and embedding head $g_{\phi}(\cdot)$ to a $d$-dimensional representation, as defined in Eq.~(\ref{eq:embedding}):
\begin{equation}
z = g_{\phi}(f_{\theta}(x)) \in \mathbb{R}^{d}
\label{eq:embedding}
\end{equation}
which is $\ell_2$-normalized according to Eq.~(\ref{eq:normalization}) used for distance-based retrieval:
\begin{equation}
\hat{z} = \frac{z}{\|z\|_2}
\label{eq:normalization}
\end{equation}

The objective is to learn parameters $(\theta,\phi)$ that produce identity-discriminative yet domain-invariant embeddings, such that samples of the same identity remain close while different identities are well separated under unseen domain shifts. To this end, we introduce an adaptive intermediate-domain learning framework that synthesizes bridge distributions across multiple source domains and dynamically regulates training through feedback-driven optimization.

\subsection{Overview of AIDA/SF-MIDA Framework}
\label{sec:aida_overview}

\begin{figure}
    \centering
    \includegraphics[width=0.8\textwidth]{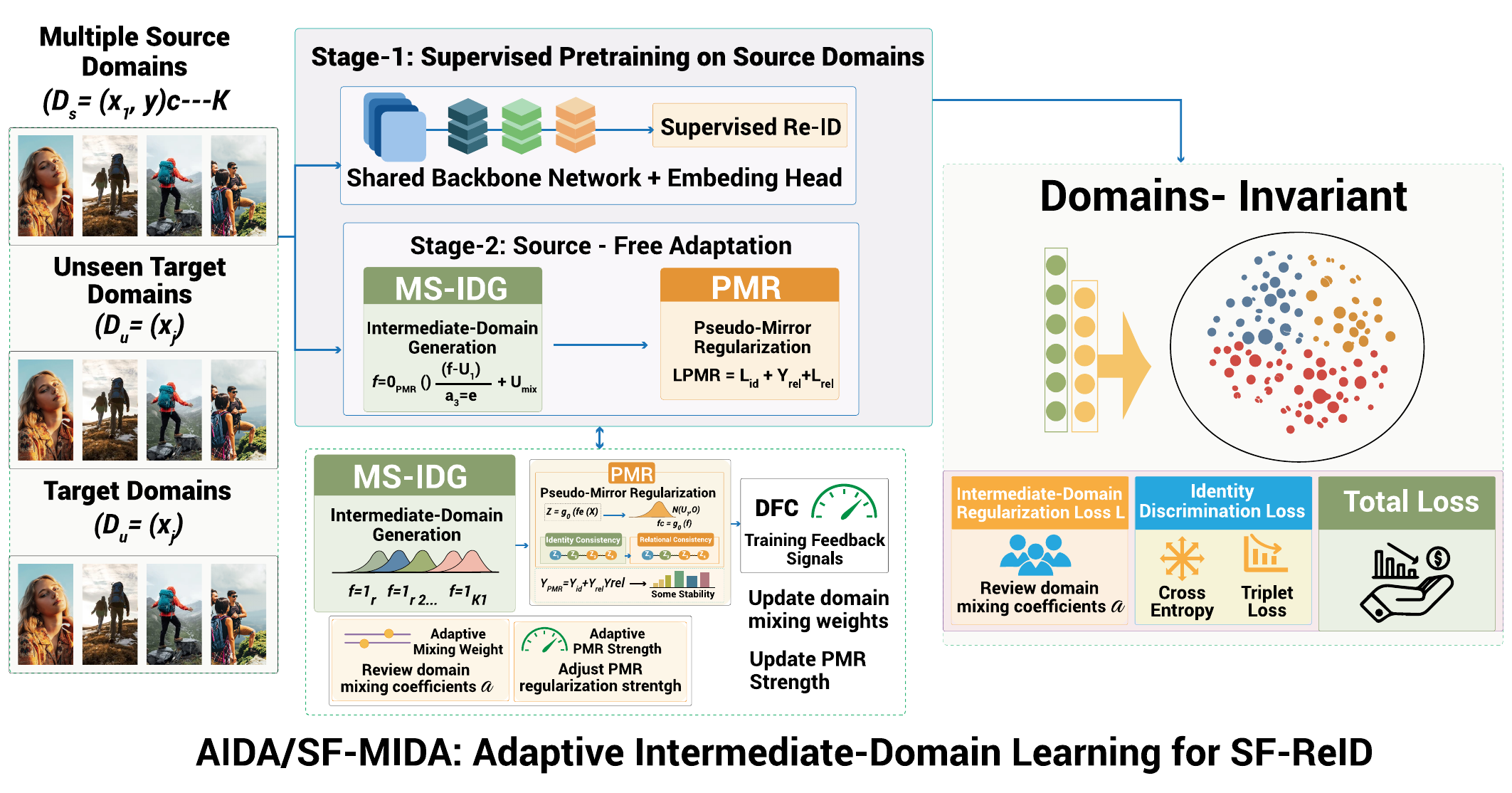}
    \caption{Overview of the proposed Adaptive Intermediate-Domain Learning framework (AIDA / SF-MIDA).}
    \label{fig:aida_overview}
\end{figure}

The AIDA/SF-MIDA framework is designed to improve the generalization ability of person re-identification models under domain shift with particular emphasis on source-free deployment. As illustrated in Fig.~\ref{fig:aida_overview} instead of adapting to a specific target domain, the framework learns identity-discriminative representations that remain robust across unseen environments by constructing adaptive intermediate domains and regulating learning through feedback-driven control. The framework follows a two-stage learning paradigm. In Stage~1, a backbone network is trained on multiple labeled source domains using standard supervised Re-ID objectives to establish strong identity-aware representations. In Stage~2, the model is refined through adaptive intermediate-domain learning, where robustness to unseen domains is enhanced without requiring access to target-domain data or source-domain samples during deployment.

AIDA/SF-MIDA integrates three tightly coupled components that operate jointly during training as shown in Fig.~\ref{fig:aida_overview}. The MS-IDG constructs adaptive intermediate representations by combining feature statistics from multiple source domains, thereby expanding the effective training distribution beyond individual sources. PMR enforces identity-level consistency under intermediate-domain perturbations stabilizing feature learning and preventing semantic drift. The DFC continuously monitors training dynamics and adaptively regulates intermediate-domain interpolation strength and regularization weights based on prediction uncertainty and optimization stability. Through the coordinated interaction of MS-IDG, PMR and DFC, AIDA/SF-MIDA enables stable optimization under domain shift while preserving identity discrimination. During inference only the backbone network and embedding head are retained ensuring compatibility with both domain generalization and source-free deployment scenarios.

\subsection{Backbone and Supervised Re-ID Learning on Sources}
\label{sec:backbone_supervised}

We adopt a standard deep embedding-based person Re-ID architecture to learn identity-discriminative representations from multiple labeled source domains. The backbone network $f_{\theta}(\cdot)$ is instantiated using a convolutional architecture pre-trained on ImageNet, followed by an embedding head $g_{\phi}(\cdot)$ that maps extracted features into a compact identity embedding space.

Given an input image $x_i^k$ sampled from the $k$-th source domain, the network produces a $d$-dimensional embedding as defined in Eq.~(\ref{eq:source_embedding}):
\begin{equation}
z_i^k = g_{\phi}\!\left(f_{\theta}(x_i^k)\right) \in \mathbb{R}^{d}
\label{eq:source_embedding}
\end{equation}
which is $\ell_2$-normalized according to Eq.~(\ref{eq:source_normalization}) to facilitate distance-based retrieval:
\begin{equation}
\hat{z}_i^k = \frac{z_i^k}{\|z_i^k\|_2}
\label{eq:source_normalization}
\end{equation}
All identity comparisons are subsequently performed in this normalized embedding space.

Supervised learning on source domains is formulated as a joint optimization of identity classification and metric learning objectives. The identity classification loss is defined as shown in Eq.~(\ref{eq:id_loss}):
\begin{equation}
\mathcal{L}_{\mathrm{id}} =
- \mathbb{E}_{(x_i^k, y_i^k)\sim \mathcal{D}_s^k}
\left[
\log p_{\theta}(y_i^k \mid x_i^k)
\right]
\label{eq:id_loss}
\end{equation}
which promotes discriminative separation among identities within each source domain.

To further enforce intra-identity compactness and inter-identity separation, we incorporate the batch-hard triplet loss, as defined in Eq.~(\ref{eq:triplet_loss}):
\begin{equation}
\mathcal{L}_{\mathrm{tri}} =
\mathbb{E}
\left[
\max\!\left(
0,\;
d(\hat{z}_i^k,\hat{z}_p^k)
-
d(\hat{z}_i^k,\hat{z}_n^k)
+ m
\right)
\right]
\label{eq:triplet_loss}
\end{equation}
where $(i,p,n)$ denote anchor positive and negative samples respectively $d(\cdot,\cdot)$ is the Euclidean distance and $m$ is a margin parameter.

The overall supervised objective for source-domain training is given in Eq.~(\ref{eq:supervised_loss}):
\begin{equation}
\mathcal{L}_{\mathrm{sup}} =
\mathcal{L}_{\mathrm{id}} + \lambda_{\mathrm{tri}} \mathcal{L}_{\mathrm{tri}}
\label{eq:supervised_loss}
\end{equation}
which establishes a strong identity-aware embedding space that serves as the foundation for subsequent adaptive intermediate-domain learning.

\subsection{Multi-Source Intermediate-Domain Generator}
\label{sec:msidg}

The Multi-Source Intermediate-Domain Generator mitigates domain bias by synthesizing intermediate representations that span the distributional space among multiple source domains. MS-IDG constructs intermediate domains using adaptive mixtures of source-domain statistics exposing the model to diverse appearance variations and reducing overfitting to individual sources.

Let $\mathcal{X}^k \sim P_k(X)$ denote the image distribution of the $k$-th source domain. An intermediate-domain distribution is defined as a convex combination of source-domain distributions, as given in Eq.~(\ref{eq:msidg_distribution}):
\begin{equation}
P_{\text{int}}(X) = \sum_{k=1}^{K} \alpha_k \, P_k(X)
\label{eq:msidg_distribution}
\end{equation}
where the mixing coefficients $\boldsymbol{\alpha} = [\alpha_1,\dots,\alpha_K]$ satisfy the simplex constraints in Eq.~(\ref{eq:alpha_constraint}):
\begin{equation}
\alpha_k \geq 0, \quad \sum_{k=1}^{K} \alpha_k = 1
\label{eq:alpha_constraint}
\end{equation}

Since direct sampling from $P_{\text{int}}(X)$ is intractable, MS-IDG operates in the feature-statistics space. Given samples $x^a \sim P_a(X)$ and $x^b \sim P_b(X)$ from distinct source domains, feature maps are extracted using the backbone network, as defined in Eq.~(\ref{eq:feature_extraction}):
\begin{equation}
f^a = f_{\theta}(x^a), \quad f^b = f_{\theta}(x^b)
\label{eq:feature_extraction}
\end{equation}
where $f_{\theta}(\cdot)$ denotes the backbone feature extractor and $\mu(\cdot)$ and $\sigma(\cdot)$ represent the channel-wise mean and standard deviation.

An intermediate feature representation is constructed via statistics transfer, as defined in Eq.~(\ref{eq:feature_transfer}):
\begin{equation}
\tilde{f} =
\sigma(f^b)
\cdot
\frac{f^a - \mu(f^a)}{\sigma(f^a)}
+
\mu(f^b)
\label{eq:feature_transfer}
\end{equation}
which preserves identity-related content from $x^a$ while injecting domain-level statistics from $x^b$.

For the multi-source setting, this operation is generalized by aggregating statistics from all source domains. The resulting intermediate representation is defined in Eq.~(\ref{eq:msidg_multi}):
\begin{equation}
\tilde{f} =
\sum_{k=1}^{K}
\alpha_k
\left(
\sigma(f^k)
\cdot
\frac{f - \mu(f)}{\sigma(f)}
+
\mu(f^k)
\right)
\label{eq:msidg_multi}
\end{equation}
where $f$ denotes the feature representation of an anchor sample and $\{f^k\}$ correspond to feature maps drawn from each source domain. This formulation enables smooth interpolation across multiple source distributions while avoiding over-specialization to any single domain.

The intermediate representation is subsequently mapped into the embedding space. As defined in Eq.~(\ref{eq:msidg_embedding}), the normalized intermediate embedding is obtained as
\begin{equation}
\tilde{z} = \frac{g_{\phi}(\tilde{f})}{\|\;g_{\phi}(\tilde{f})\;\|_2}
\label{eq:msidg_embedding}
\end{equation}
and is used for regularization and feedback-driven adaptation in subsequent learning stages. By explicitly modeling intermediate domains through convex combinations of source-domain statistics, MS-IDG expands the effective training distribution and encourages the learning of identity representations that are less sensitive to domain-specific appearance variations.

\subsection{Pseudo-Mirror Regularization}
\label{sec:pmr}

Pseudo-Mirror Regularization is designed to preserve identity structure under intermediate-domain transformations. As intermediate-domain generation introduces controlled distributional perturbations, PMR constrains the embedding space to remain invariant to such changes, thereby preventing semantic drift and identity collapse during training.

Let $x \sim P_k(X)$ denote a source-domain sample and $\hat{x}=\mathcal{M}(x)$ its pseudo-mirrored counterpart generated via domain-statistics perturbation. Given the embedding function $z(\cdot)=g_{\phi}(f_{\theta}(\cdot))$, PMR enforces first-order identity consistency by minimizing discrepancies between embeddings of original and pseudo-mirrored samples, as defined in Eq.~(\ref{eq:lpmr}):
\begin{equation}
\mathcal{L}_{\mathrm{pmr}} =
\mathbb{E}_{x}
\left[
\left\|
\hat{z}(x) - \hat{z}(\hat{x})
\right\|_2^2
\right]
\label{eq:lpmr}
\end{equation}
where $\hat{z}(\cdot)$ denotes $\ell_2$-normalized embeddings. This loss encourages identity-preserving representations that remain stable under domain-level appearance variations.

Beyond pointwise consistency, PMR further preserves the relational structure of the embedding space. Let $\mathcal{P}$ denote a set of sample pairs drawn from a mini-batch. For each pair $(i,j)\in\mathcal{P}$, relational consistency is enforced between pairwise distances computed from the original and pseudo-mirrored embeddings, as defined in Eq.~(\ref{eq:lrel}):

\begin{equation}
\mathcal{L}_{\mathrm{rel}} =
\mathbb{E}_{(i,j)\in\mathcal{P}}
\left[
\left|
d(\hat{z}_i,\hat{z}_j)
-
d(\hat{z}_i^{\,\prime},\hat{z}_j^{\,\prime})
\right|
\right]
\label{eq:lrel}
\end{equation}
where $\hat{z}_i^{\,\prime}$ and $\hat{z}_j^{\,\prime}$ denote embeddings of pseudo-mirrored samples and $d(\cdot,\cdot)$ is a distance metric. This constraint preserves local neighborhood geometry under intermediate-domain perturbations.

The overall PMR objective integrates identity-level and relational consistency terms, as defined in Eq.~(\ref{eq:lpmr_final}):
\begin{equation}
\mathcal{L}_{\mathrm{PMR}} =
\mathcal{L}_{\mathrm{pmr}}
+
\lambda_{\mathrm{rel}}
\mathcal{L}_{\mathrm{rel}}
\label{eq:lpmr_final}
\end{equation}
where $\lambda_{\mathrm{rel}}$ controls the influence of relational preservation. Together, these constraints enhance robustness to domain shifts while maintaining discriminative identity structure in the embedding space.

\subsection{Dynamic Feedback Controller}
\label{sec:dfc}

The Dynamic Feedback Controller regulates the strength of intermediate-domain generation and regularization during training. While adaptive intermediate-domain learning expands domain coverage uncontrolled perturbations may lead to insufficient adaptation or excessive distortion of identity features. DFC addresses this issue through a closed-loop control mechanism that dynamically adjusts key training variables based on internal feedback signals, thereby stabilizing optimization.

To characterize the training state, two complementary indicators are employed. First, prediction uncertainty on intermediate-domain samples is quantified using batch-wise entropy, as defined in Eq.~(\ref{eq:entropy}):

\begin{equation}
\mathcal{E} =
- \frac{1}{B}
\sum_{i=1}^{B}
\sum_{c}
p_{i,c}
\log p_{i,c}
\label{eq:entropy}
\end{equation}
where $B$ denotes the mini-batch size and $p_{i,c}$ represents the predicted posterior probability of class $c$ for the $i$-th sample. Elevated entropy indicates ambiguous identity predictions.

Second, optimization stability is assessed through the variance of loss gradients with respect to model parameters, as defined in Eq.~(\ref{eq:gradvar}):

\begin{equation}
\mathcal{V} =
\mathrm{Var}\!\left(
\nabla_{\theta} \mathcal{L}_{\mathrm{total}}
\right)
\label{eq:gradvar}
\end{equation}
where $\mathcal{L}_{\mathrm{total}}$ denotes the overall training objective. Large gradient variance reflects unstable parameter updates.

Based on these feedback signals, DFC dynamically updates the multi-source mixing weights
$\boldsymbol{\alpha}=[\alpha_1,\ldots,\alpha_K]$, which control the contribution of each source domain to the intermediate distribution. The update rule is given by Eq.~(\ref{eq:alpha_update}):

\begin{equation}
\boldsymbol{\alpha} \leftarrow
\Pi_{\Delta}
\left(
\boldsymbol{\alpha}
-
\eta_{\alpha}
\left(
\frac{\mathcal{E}}{\mathcal{E}_{\max}}
+
\frac{\mathcal{V}}{\mathcal{V}_{\max}}
\right)\mathbf{1}
\right)
\label{eq:alpha_update}
\end{equation}
where $\Pi_{\Delta}(\cdot)$ projects the updated weights onto the probability simplex, $\mathcal{E}_{\max}$ and $\mathcal{V}_{\max}$ are normalization constants estimated from running statistics and $\eta_{\alpha}$ controls the controller step size.

In parallel, the regularization strength of Pseudo-Mirror Regularization is adaptively adjusted according to Eq.~(\ref{eq:lambda_update}):
\begin{equation}
\lambda_{\mathrm{PMR}} \leftarrow
\mathrm{clip}
\left(
\lambda_{\mathrm{PMR}}
+
\eta_{\lambda}
\left(
\frac{\mathcal{E}}{\mathcal{E}_{\max}}
+
\frac{\mathcal{V}}{\mathcal{V}_{\max}}
\right),
\, 0,\, \lambda_{\max}
\right)
\label{eq:lambda_update}
\end{equation}
where $\eta_{\lambda}$ controls the adaptation rate and $\lambda_{\max}$ bounds the maximum regularization strength.

\subsection{Optimization Procedure}
\label{sec:optimization}

\begin{figure}
  \centering
  \includegraphics[width=0.75\linewidth]{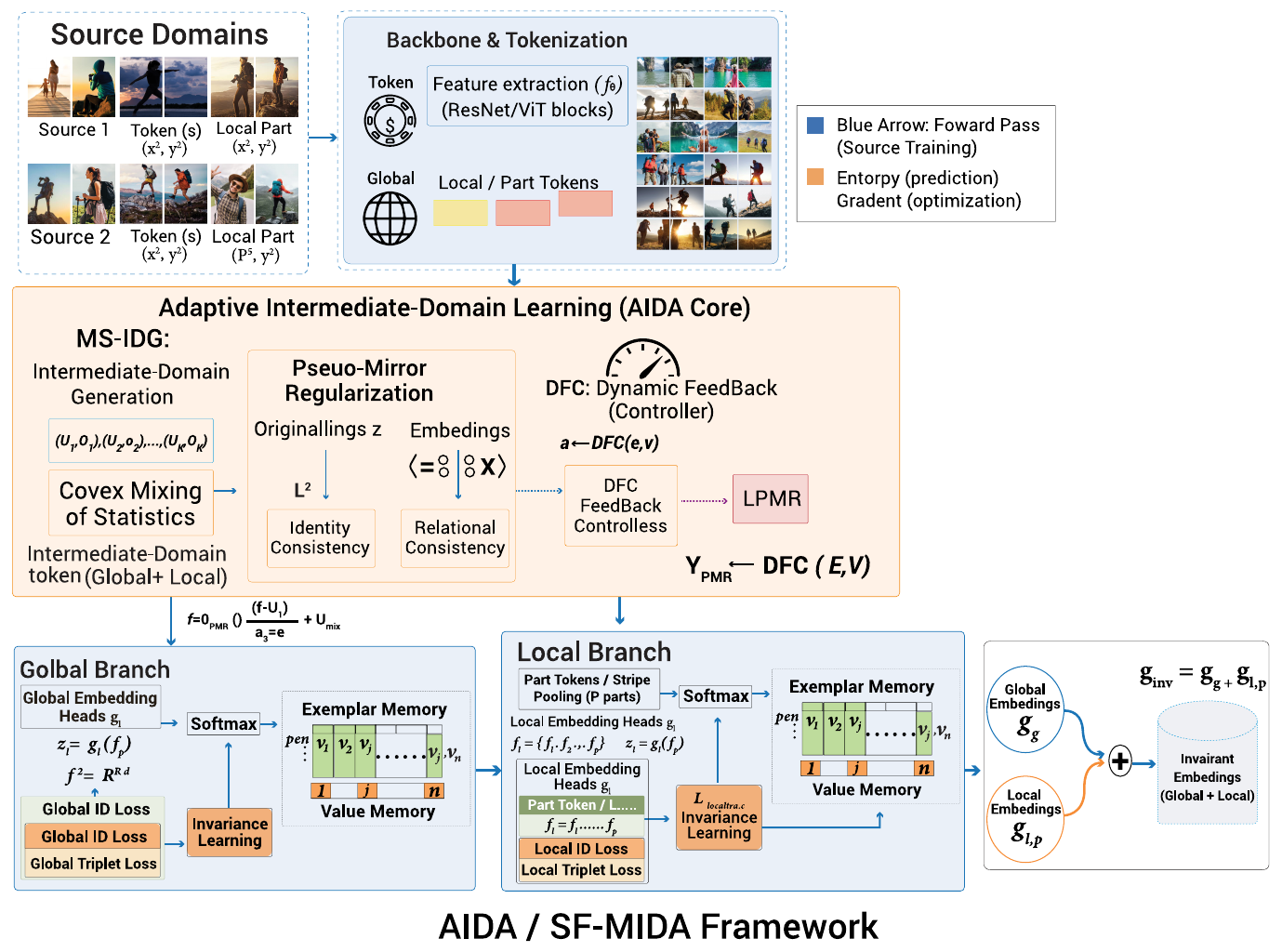}
  \caption{Overview of the proposed AIDA / SF-MIDA framework. Labeled source domains are processed by a shared backbone to extract global and local representations. The AIDA core consists of three modules: Multi-Source Intermediate Domain Generation (MS-IDG), Pseudo-Mirror Regularization (PMR), and a Dynamic Feedback Controller (DFC). These components jointly construct adaptive intermediate domains and regulate training dynamics to produce domain-invariant embeddings for person re-identification.}
  \label{fig:aida_architecture}
\end{figure}

The AIDA/SF-MIDA framework is trained through an iterative optimization process that jointly updates network parameters and controller variables. The backbone and embedding head are optimized using standard gradient-based methods, while the dynamic feedback controller (DFC) operates as a closed-loop mechanism that adaptively regulates intermediate-domain generation and regularization strength based on training feedback. Let $(\theta,\phi)$ denote the parameters of the backbone and embedding head, and $\Omega=\{\boldsymbol{\alpha},\lambda_{\mathrm{PMR}}\}$ represent the controller state. Training proceeds over multiple epochs using mini-batches sampled from labeled source domains following the optimization pipeline illustrated in Fig.~\ref{fig:aida_architecture}.

At each iteration, supervised identity learning, intermediate-domain consistency and pseudo-mirror regularization are jointly optimized under a unified training objective. Feedback indicators derived from prediction uncertainty and optimization stability are concurrently used to update the controller state $\Omega$. By decoupling controller updates from gradient-based parameter optimization, the proposed framework adaptively adjusts intermediate-domain perturbation strength without destabilizing network training, ensuring stable convergence and improved robustness to unseen domain shifts. The overall training procedure of AIDA/SF-MIDA is summarized in Algorithm~\ref{alg:aida}.

\begin{algorithm}[!htbp]
\caption{AIDA / SF-MIDA: Adaptive Intermediate-Domain Learning for Source-Free Person Re-Identification}
\label{alg:aida}
\scriptsize
\DontPrintSemicolon
\SetKwInOut{Require}{Require}
\SetKwInOut{Ensure}{Ensure}
\SetKw{KwRet}{Return}

\Require{Source sets $\{\mathcal{D}_s^k\}_{k=1}^{K}$; optional target set $\mathcal{D}_u$; backbone $f_\theta$; classifier head $g_\phi$; controller $\Omega=\{\boldsymbol{\alpha},\lambda_{\mathrm{PMR}}\}$; epochs $E_{\mathrm{sup}}$ and $E_{\mathrm{aida}}$}
\Ensure{Trained parameters $(\theta,\phi)$}

Initialize $(\theta,\phi)$, $\boldsymbol{\alpha}\leftarrow \frac{1}{K}\mathbf{1}$, and $\lambda_{\mathrm{PMR}}\leftarrow\lambda_0$\;

\BlankLine
\textbf{Stage 1: Supervised pre-training}\;
\For{$e \leftarrow 1$ \KwTo $E_{\mathrm{sup}}$}{
  \ForEach{mini-batch $\{(x^k,y^k)\}$ sampled from $\{\mathcal{D}_s^k\}$}{
    $f^k \leftarrow f_\theta(x^k)$; $z^k \leftarrow g_\phi(f^k)$; $\hat{z}^k \leftarrow \mathrm{norm}(z^k)$\;
    $\mathcal{L}_{\mathrm{sup}} \leftarrow \mathcal{L}_{\mathrm{id}}(\hat{z}^k,y^k)+\lambda_{\mathrm{tri}}\mathcal{L}_{\mathrm{tri}}(\hat{z}^k,y^k)$\;
    Update $(\theta,\phi)$ using $\mathcal{L}_{\mathrm{sup}}$\;
  }
}

\BlankLine
\textbf{Stage 2: AIDA training}\;
\For{$e \leftarrow 1$ \KwTo $E_{\mathrm{aida}}$}{
  \ForEach{mini-batch $\{(x^k,y^k)\}$ sampled from $\{\mathcal{D}_s^k\}$}{
    $f \leftarrow f_\theta(x)$; $\hat{z} \leftarrow \mathrm{norm}(g_\phi(f))$\;
    Compute $\{(\mu^k,\sigma^k)\}_{k=1}^{K}$ from $\{f^k\}$\;
    $(\bar{\mu},\bar{\sigma}) \leftarrow \sum_k \alpha_k(\mu^k,\sigma^k)$\;
    $\tilde{f} \leftarrow \bar{\sigma}\cdot\dfrac{f-\mu(f)}{\sigma(f)+\epsilon}+\bar{\mu}$\;
    $\hat{\tilde{z}} \leftarrow \mathrm{norm}(g_\phi(\tilde{f}))$\;
    $\mathcal{L}_{\mathrm{PMR}} \leftarrow \mathcal{L}_{\mathrm{pmr}}(\hat{z},\hat{\tilde{z}})+\lambda_{\mathrm{rel}}\mathcal{L}_{\mathrm{rel}}(\hat{z},\hat{\tilde{z}})$\;
    $\mathcal{L}_{\mathrm{total}} \leftarrow \mathcal{L}_{\mathrm{sup}}+\lambda_{\mathrm{PMR}}\mathcal{L}_{\mathrm{PMR}}$\;
    Update $(\theta,\phi)$ using $\mathcal{L}_{\mathrm{total}}$\;
    Compute entropy $\mathcal{E}$ and gradient variance $\mathcal{V}$\;
    Update $\boldsymbol{\alpha}$ and $\lambda_{\mathrm{PMR}}$ using DFC feedback\;
  }
}

\BlankLine
\If{$\mathcal{D}_u$ is available}{
  \ForEach{target mini-batch $\{x^u\}$}{
    $\hat{z}^u \leftarrow \mathrm{norm}(g_\phi(f_\theta(x^u)))$\;
    Apply PMR-style consistency and update $(\theta,\phi)$\;
    Update $\Omega$ using DFC feedback\;
  }
}

\KwRet{$(\theta,\phi)$}\;
\end{algorithm}

\section{Experiments}
\label{sec:experiments}

\subsection{Datasets and Evaluation Protocol}
\label{sec:datasets_protocol}
\subsubsection{Datasets}
We evaluate the proposed AIDA/SF-MIDA framework on five widely used person re-identification benchmarks including Market-1501~\cite{umair2026unifying}, DukeMTMC-ReID~\cite{xiao2025reid}, MSMT17~\cite{tian2025part}, CUHK03~\cite{chen2025improved} and PersonX~\cite{peng2024adapt}. These datasets exhibit significant variations in camera viewpoints, illumination conditions, background clutter and acquisition environments, providing diverse evaluation scenarios for domain generalization and adaptation. Dataset statistics are summarized in Table~\ref{tab:dataset_stats}. Market-1501 and DukeMTMC-ReID are captured under relatively controlled environments and DukeMTMC-ReID is used strictly for academic research following standard person re-identification evaluation protocols. MSMT17 introduces larger domain shifts due to diverse lighting and weather conditions. CUHK03 presents additional challenges caused by detection noise, while PersonX is a synthetic dataset designed to evaluate generalization under controlled domain shifts. CUHK03 and PersonX are primarily used for extended domain generalization analysis and are not included in the main UDA benchmark comparisons unless otherwise stated.

\subsubsection{Evaluation Metrics}
Following standard practice in person re-identification, we report the Cumulative Matching Characteristic (CMC) at Rank-$k$ ($k=\{1,5,10\}$) and mean Average Precision (mAP) to evaluate retrieval performance. For representation-level analysis under domain generalization and source-free settings normalized mutual information (NMI) and silhouette score are additionally reported when applicable. 

\subsubsection{Evaluation Protocol}
We consider three evaluation settings. In the unsupervised domain adaptation setting, labeled source data and unlabeled target data are jointly used for adaptation. In the domain generalization setting, models are trained on multiple labeled source domains and evaluated on an unseen target domain following a leave-one-domain-out protocol. In the source-free setting (SF-MIDA), source data are not accessible during adaptation only the pretrained model parameters are retained and refinement is performed using unlabeled samples from the target domain when available, following standard source-free adaptation practice without accessing any source-domain images.

\begin{table}[!htbp]
\centering
\caption{Statistics of datasets used in our experiments.}
\label{tab:dataset_stats}

\footnotesize
\setlength{\tabcolsep}{2pt}
\renewcommand{\arraystretch}{.90}

\begin{tabular}{lcccc}
\toprule
Dataset & \#Images & \#IDs & \#Cameras & Type \\
\midrule
Market-1501     & 32,668  & 1,501 & 6  & Real \\
DukeMTMC-ReID   & 36,411  & 1,404 & 8  & Real \\
MSMT17          & 126,441 & 4,101 & 15 & Real \\
CUHK03          & 13,164  & 1,467 & 2  & Real \\
PersonX         & 273,456 & 1,266 & 6  & Synthetic \\
\bottomrule
\end{tabular}

\end{table}
\subsection{Implementation Details}
\label{sec:implementation}

We adopt ResNet-50 pre-trained on ImageNet as the backbone feature extractor. Input images are resized to $256\times128$. The embedding head follows the standard Re-ID design, while AIDA/SF-MIDA augments the backbone with MS-IDG, PMR and DFC to regulate intermediate-domain construction and training stability. 
For single-source UDA, we follow the IDM++~\cite{dai2025idmpp} training pipeline using DSBN and global average pooling for fair comparison with prior methods. Domain generalization and source-free settings employ the full AIDA architecture with GeM pooling and Sharpness-Aware Minimization (SAM). 
Models are optimized using Adam with an initial learning rate of $3.5\times10^{-4}$ for UDA and $3.0\times10^{-4}$ for DG pre-training, followed by source-free refinement at $1.0\times10^{-4}$. Training lasts 60 epochs for UDA, 50 epochs for DG pre-training and 30 epochs for source-free refinement. Standard augmentations including random horizontal flipping and random erasing are applied.

\subsection{Experimental Phases Overview}
\label{sec:phases}
We organize the experimental evaluation into four phases aligned with practical deployment settings: (i) multi-source domain generalization under a leave-one-domain-out protocol; (ii) source-free adaptation using a two-stage pretraining and adaptation pipeline; (iii) single-source unsupervised domain adaptation across standard cross-domain Re-ID benchmarks and (iv) ablation and efficiency analyses to validate the individual contributions and computational cost of MS-IDG, PMR and DFC.

\FloatBarrier
\section{Results and Analysis}
\label{sec:results}

\subsection{Multi-Source Domain Generalization Results}
\label{sec:results_dg}

We evaluate the proposed AIDA framework under a multi-source domain generalization setting using a leave-one-domain-out protocol. In this setting, models are trained on multiple labeled source datasets and directly evaluated on an unseen target domain without accessing any target-domain data during training. This protocol reflects realistic deployment scenarios in which the target environment is unknown in advance.

Table~\ref{tab:dg_main} reports Rank-$k$ accuracy, mean Average Precision (mAP) and clustering-based indicators (NMI and Silhouette) to assess both retrieval performance and the quality of the learned feature space. The results show that AIDA achieves stable and competitive generalization across heterogeneous unseen targets with varying levels of domain shift, camera configurations and dataset scale. The performance trends observed across different targets highlight the effectiveness of adaptive intermediate-domain learning in mitigating domain bias while preserving identity discrimination under severe distribution shifts.

\begin{table}[!htbp]
\centering
\caption{Multi-source domain generalization (DG) results under leave-one-out evaluation. Rank-1/5/10, mAP and clustering indicators are reported.}
\label{tab:dg_main}

\footnotesize
\setlength{\tabcolsep}{2pt}
\renewcommand{\arraystretch}{.95}

\begin{tabular}{l l c c c c c c}
\toprule
\textbf{Train Sources} & \textbf{Test Target} & \textbf{R1} & \textbf{R5} & \textbf{R10} & \textbf{mAP} & \textbf{NMI} & \textbf{Sil.} \\
\midrule
Market + Duke & MSMT17 & 57.8 & 58.1 & 59.2 & 39.9 & 0.742 & 0.312 \\
Market + MSMT17 & Duke & 78.5 & 88.2 & 91.7 & 64.3 & 0.768 & 0.328 \\
Duke + MSMT17 & Market & 89.7 & 95.4 & 97.2 & 76.8 & 0.791 & 0.345 \\
Market + Duke + MSMT17 & CUHK03 & 51.3 & 52.7 & 53.8 & 49.9 & 0.756 & 0.321 \\
\midrule
\textbf{Average} & -- & \textbf{69.3} & \textbf{73.6} & \textbf{75.5} & \textbf{57.7} & \textbf{0.764} & \textbf{0.327} \\
\bottomrule
\end{tabular}
\end{table}

As shown in Table~\ref{tab:dg_main}, MSMT17 remains the most challenging unseen target due to its large scale and highly diverse acquisition conditions, whereas Market-1501 exhibits stronger generalization when trained using complementary source domains. AIDA achieves 76.8 mAP on Market-1501 when trained on DukeMTMC-ReID and MSMT17 demonstrating the benefit of multi-source intermediate-domain exposure. Performance on CUHK03 reaches 49.9 mAP, demonstrating robustness under severe detection noise and limited camera diversity. Similar robustness trends are also observed on the synthetic PersonX benchmark as discussed in Section~\ref{sec:personx} further supporting the effectiveness of adaptive intermediate-domain learning under diverse and unseen domain shifts.

\FloatBarrier
\subsection{Source-Free Adaptation Results}
\label{sec:results_sf}

We evaluate SF-MIDA under a source-free adaptation setting using a two-stage training protocol. In the first stage, the model is pre-trained on multiple labeled source domains, namely Market-1501, DukeMTMC-ReID and MSMT17. In the second stage, target-only refinement is performed using unlabeled samples from the target domain without accessing any source-domain images. This setting reflects realistic deployment scenarios in which source data cannot be retained due to privacy or storage constraints. During this refinement phase the model relies on pseudo-label estimation and consistency-driven optimization to progressively adapt the learned representation to the target distribution. Importantly, only the learned parameters are transferred from the pre-training stage, ensuring that no source-domain samples or statistics are reused during adaptation.

Table~\ref{tab:sf_main} reports Rank-$k$ accuracy, mAP and clustering-based indicators for each target domain. The results show that SF-MIDA maintains strong retrieval performance under strict source-free constraints. High performance is achieved on Market-1501 and DukeMTMC-ReID indicating that multi-source intermediate-domain pre-training provides a robust initialization for subsequent target-only refinement. Although MSMT17 remains the most challenging target due to its scale and environmental diversity SF-MIDA still achieves competitive performance confirming the robustness of the proposed framework under severe domain shifts. SF-MIDA attains an average mAP of 76.9 across the three target domains under the source-free adaptation protocol.

\begin{table}[!htbp]
\centering
\caption{Source-free adaptation (SF-MIDA) results (target-only refinement). Rank-$k$, mAP and clustering indicators are reported.}
\label{tab:sf_main}
\footnotesize
\setlength{\tabcolsep}{2pt}
\renewcommand{\arraystretch}{.95}
\begin{tabular}{l c c c c c c}
\toprule
\textbf{Target} & \textbf{R1} & \textbf{R5} & \textbf{R10} & \textbf{mAP} & \textbf{NMI} & \textbf{Sil.} \\
\midrule
Market-1501 & 97.2 & 98.1 & 99.2 & 86.4 & 0.784 & 0.338 \\
DukeMTMC-ReID & 95.4 & 96.2 & 97.4 & 84.8 & 0.761 & 0.325 \\
MSMT17 & 69.1 & 73.4 & 77.6 & 59.5 & 0.728 & 0.301 \\
\midrule
\textbf{Average} & \textbf{87.2} & \textbf{89.2} & \textbf{91.4} & \textbf{76.9} & \textbf{0.758} & \textbf{0.321} \\
\bottomrule
\end{tabular}
\end{table}

As reported in Table~\ref{tab:sf_main}, SF-MIDA maintains strong and stable performance across all evaluated target domains despite the absence of source-domain images during target refinement. The highest performance is observed on Market-1501 (86.4 mAP) followed by DukeMTMC-ReID (84.8 mAP), while MSMT17 remains the most challenging target with 59.5 mAP due to its large scale and complex environmental variations. These results indicate that multi-source intermediate-domain pre-training provides a robust initialization for target-only adaptation enabling the proposed framework to retain strong discriminative capability under strict source-free deployment constraints.

\FloatBarrier
\subsection{Single-Source UDA Results}
\label{sec:results_uda}

We report single-source UDA results for person Re-ID under standard cross-domain evaluation protocols including Market 1501$\leftrightarrow$DukeMTMC-ReID and transfer settings involving MSMT17. Performance is evaluated using Rank-1 (R1) accuracy and mAP which are the primary metrics adopted in recent UDA Re-ID literature. Table~\ref{tab:uda_main} summarizes the cross-domain adaptation performance of the proposed method under single-source settings.

The results indicate strong adaptation capability between Market-1501 and DukeMTMC-ReID where relatively smaller domain gaps exist. Transfers involving MSMT17 remain more challenging due to its large scale and diverse acquisition conditions. Even so, the proposed method maintains competitive performance across all evaluated transfer directions.

\begin{table}[!htbp]
\centering
\caption{Single-source unsupervised domain adaptation (UDA) results for person
re-identification using our method. Rank-1 (R1) accuracy and mean Average
Precision (mAP) are reported.}
\label{tab:uda_main}

\footnotesize
\setlength{\tabcolsep}{2pt}
\renewcommand{\arraystretch}{0.90}

\begin{adjustbox}{max width=\columnwidth}
\begin{tabular}{l c c}
\toprule
\textbf{Transfer (Source $\rightarrow$ Target)} & \textbf{R1} & \textbf{mAP} \\
\midrule
Market-1501 $\rightarrow$ DukeMTMC-reID & 92.5 & 84.9 \\
DukeMTMC-reID $\rightarrow$ Market-1501 & 95.3 & 89.2 \\
Market-1501 $\rightarrow$ MSMT17 & 72.1 & 58.9 \\
DukeMTMC-reID $\rightarrow$ MSMT17 & 73.5 & 46.1 \\
MSMT17 $\rightarrow$ DukeMTMC-reID & 86.7 & 75.9 \\
MSMT17 $\rightarrow$ Market-1501 & 96.8 & 89.6 \\
\bottomrule
\end{tabular}
\end{adjustbox}
\end{table}

\FloatBarrier

\subsection{Comparison with State-of-the-Art}
\label{sec:sota}

We compare the proposed method with representative recent person Re-ID approaches under two evaluation settings: domain generalization and source-free adaptation. As most existing methods focus on only one setting, DG and SF results are reported separately in Tables~\ref{tab:sota_dg_final} and~\ref{tab:sota_sf_refined}. Under the multi-source DG protocol, AIDA achieves strong generalization across unseen targets and attains the highest mAP on MSMT17 using a consistent ResNet-50 backbone indicating improved robustness under severe domain shifts. Competitive performance is also observed on Market-1501, DukeMTMC-ReID and CUHK03. Under strict source-free constraints SF-MIDA remains competitive across all evaluated targets without accessing any source-domain data demonstrating the effectiveness of the proposed framework for practical deployment.

\begin{table}[!htbp]
\centering
\caption{Comparison with DG-ReID methods under leave-one-domain-out evaluation.
Rank-1/5/10 accuracy and mAP are reported when available.}
\label{tab:sota_dg_final}
\footnotesize
\setlength{\tabcolsep}{2pt}
\renewcommand{\arraystretch}{.95}
\begin{adjustbox}{max width=\textwidth}
\begin{tabular}{l *{16}{c}}
\toprule
\multirow{2}{*}{Method}
& \multicolumn{4}{c}{Market-1501}
& \multicolumn{4}{c}{DukeMTMC}
& \multicolumn{4}{c}{MSMT17}
& \multicolumn{4}{c}{CUHK03} \\
\cmidrule(lr){2-5} \cmidrule(lr){6-9} \cmidrule(lr){10-13} \cmidrule(lr){14-17}
& R1 & R5 & R10 & mAP
& R1 & R5 & R10 & mAP
& R1 & R5 & R10 & mAP
& R1 & R5 & R10 & mAP \\
\midrule
MixNorm \cite{qi2022mixnormalization} (2022)
& 78.9 & -- & -- & 51.4
& 70.8 & -- & -- & 49.9
& 47.2 & -- & -- & 19.4
& 29.6 & -- & -- & 29.0 \\
QAConv-G \cite{liao2022graph} (2022)
& 79.1 & -- & -- & 53.8
& 72.4 & -- & -- & 54.5
& 35.9 & -- & -- & 34.2
& 46.5 & -- & -- & 17.1 \\
DCCL \cite{gong2023debiased} (2023)
& 85.3 & -- & -- & 63.2
& 75.9 & -- & -- & 60.0
& 40.3 & -- & -- & 38.7
& 47.4 & -- & -- & 20.5 \\
PAT \cite{ni2023partaware} (2023)
& 80.0 & -- & -- & 50.6
& 71.5 & -- & -- & 55.9
& 44.5 & -- & -- & 19.0
& 31.6 & -- & -- & 31.1 \\
STL \cite{wang2024adaptive} (2024)
& 85.7 & -- & -- & 63.3
& 74.6 & -- & -- & 57.1
& 45.1 & -- & -- & 21.3
& 41.9 & -- & -- & 40.4 \\
IGMG \cite{bhuiyan2024igmg} (2024)
& 83.1 & -- & -- & 57.1
& 76.2 & -- & -- & 57.7
& 37.3 & -- & -- & 36.3
& 41.6 & -- & -- & 19.7 \\
SALDG \cite{guo2024domain} (2024)
& 88.0 & -- & -- & 67.0
& 77.6 & -- & -- & 62.2
& 49.1 & -- & -- & 20.6
& 50.9 & -- & -- & 49.5 \\
DCAC \cite{li2025unleashing} (2025)
& 80.0 & -- & -- & 56.7
& 75.4 & -- & -- & 58.9
& 56.7 & -- & -- & 27.5
& 43.6 & -- & -- & 42.5 \\
IDM++ \cite{dai2025bridging} (2025)
& 82.2 & -- & -- & 58.8
& 74.5 & -- & -- & 58.3
& 44.3 & -- & -- & 20.1
& 40.9 & -- & -- & 39.9 \\
DualNormNP \cite{bhuiyan2025optimizing} (2025)
& 85.7 & -- & -- & 61.2
& 76.8 & -- & -- & 58.9
& 39.1 & -- & -- & 38.6
& 48.9 & -- & -- & 21.5 \\
\midrule
\textbf{AIDA (Ours)}
& \textbf{89.7} & \textbf{95.4} & \textbf{97.2} & \textbf{76.8}
& \textbf{78.5} & \textbf{88.2} & \textbf{91.7} & \textbf{62.3}
& \textbf{57.8} & \textbf{58.1} & \textbf{59.2} & \textbf{39.7}
& \textbf{51.3} & \textbf{52.7} & \textbf{53.8} & \textbf{49.9} \\
\bottomrule
\end{tabular}
\end{adjustbox}
\end{table}

\begin{table}[!htbp]
\centering
\caption{Comparison with state-of-the-art source-free (SF) person Re-ID methods. Rank-1/5/10 accuracy and mAP are reported when available.}
\label{tab:sota_sf_refined}
\footnotesize
\setlength{\tabcolsep}{2pt}
\renewcommand{\arraystretch}{0.90}
\begin{tabular}{l c c c c c c}
\toprule
Method & Year & Target & R1 & R5 & R10 & mAP \\
\midrule
P2LR \cite{han2022probabilistic_uncertainty} & 2022 & Market & 92.6 & 97.4 & 98.3 & 81.0 \\
SECRET \cite{he2022secret} & 2022 & Duke & 93.3 & -- & -- & 83.0 \\
MDJL \cite{chen2023multi_domain_joint} & 2023 & MSMT17 & 40.3 & 51.2 & 56.3 & 17.1 \\
INCLR \cite{zha2023intensifying} & 2023 & MSMT17 &  61.2 & -- & -- & 30.1 \\
S2ADAP \cite{qu2024sfda_style} & 2024 & Duke &  83.7 & 91.3 & 94.1 & 71.8 \\
IAMT \cite{qu2024instance} & 2024 & Market &  93.6 & -- & -- & 82.8 \\
AAMT \cite{qu2024aamt} & 2024 & Duke & 85.7 & -- & -- & 74.7 \\
SecureDA \cite{qu2025secureda} & 2025 & Market & 94.5 & -- & -- & 84.2 \\
RULER \cite{zheng2025ruler} & 2025 & Market & 93.5 & 97.0 & 98.0 & 84.2\\
RULER \cite{zheng2025ruler} & 2025 & Duke & 85.4 & 92.5 & 94.8 & 74.1 \\
\midrule
\textbf{SF-MIDA (Ours)} & 2026 & Market & \textbf{97.2} & \textbf{98.1} & \textbf{99.2} & \textbf{86.4} \\
\textbf{SF-MIDA (Ours)} & 2026 & Duke & \textbf{95.4} & \textbf{96.2} & \textbf{97.4} & \textbf{84.8} \\
\textbf{SF-MIDA (Ours)} & 2026 & MSMT17 & \textbf{69.1} & \textbf{73.4} & \textbf{77.6} & \textbf{59.5} \\
\bottomrule
\end{tabular}

\par\vspace{2pt}
\footnotesize
Note: Results are reported as provided in the original papers. Differences in backbone architectures, training protocols and evaluation settings may affect direct comparability.
\end{table}

\subsection{Additional Robustness Evaluation on Synthetic Data}
\label{sec:personx}

In addition to real-world benchmarks, we evaluate the proposed framework on the synthetic PersonX dataset to assess robustness under controlled domain shifts. As an unseen synthetic target, PersonX introduces large appearance variations through rendering while preserving identity consistency providing a complementary evaluation beyond real-camera datasets. 
As shown in Table~\ref{tab:personx_eval} AIDA maintains strong generalization performance under synthetic domain shifts. Despite substantial appearance discrepancies introduced by synthetic rendering the model achieves competitive retrieval accuracy indicating improved robustness beyond real-world acquisition conditions.

\begin{table}[!htbp]
\centering
\caption{Additional robustness evaluation on the synthetic PersonX dataset.
Models are trained on Market, Duke and MSMT, and evaluated on PersonX.
(Market = Market-1501; Duke = DukeMTMC-reID; MSMT = MSMT17.)}
\label{tab:personx_eval}
\footnotesize
\setlength{\tabcolsep}{2pt}
\renewcommand{\arraystretch}{0.95}
\begin{tabular}{p{0.52\linewidth} c c c c}
\toprule
\textbf{Train Sources} & \textbf{R1} & \textbf{R5} & \textbf{R10} & \textbf{mAP} \\
\midrule
Market + Duke + MSMT $\rightarrow$ PersonX
& \textbf{82.1} & \textbf{90.5} & \textbf{93.3} & \textbf{68.4} \\
\bottomrule
\end{tabular}
\end{table}

\subsection{Ablation Studies}
\label{sec:ablation}

To quantify the contribution of each design choice in AIDA/SF-MIDA, we conduct a comprehensive ablation study focusing on the roles of the proposed MS-IDG, PMR and the DFC. All ablations are performed under the DG setting using the same training protocols as in the main experiments to ensure a fair and controlled comparison.

Following standard practice in high-tier person Re-ID literature, we evaluate module-wise ablations on representative DG transfers:
Market-1501 + DukeMTMC-reID $\rightarrow$ MSMT17, Market-1501 + MSMT17 $\rightarrow$ DukeMTMC-reID and DukeMTMC-reID + MSMT17 $\rightarrow$ Market-1501. These transfers represent distinct domain shift characteristics and allow us to examine whether the observed improvements are consistent across different unseen target domains rather than being confined to a single evaluation scenario.

In addition to progressive module ablations, we further isolate the effect of the proposed PMR strategy by comparing it against conventional pseudo-label refinement schemes. This analysis is designed to assess the robustness of identity preservation under intermediate-domain perturbations. We further analyze the behavior of the DFC to examine how feedback-driven regulation influences training dynamics and generalization performance across heterogeneous real-world domains including CUHK03.

\begin{table}[!htbp]
\centering
\caption{Complex ablation study of AIDA under DG evaluation. We progressively
activate MS-IDG, PMR, and DFC, and report Rank-1 (R1) and mAP for each
cross-domain transfer. Avg Gain is computed w.r.t.\ the IDM++ baseline.}
\label{tab:ablation_complex}
\footnotesize
\setlength{\tabcolsep}{2pt}
\renewcommand{\arraystretch}{0.90}

\resizebox{\textwidth}{!}{%
\begin{tabular}{l c c c c c c c c c c}
\toprule
\multirow{2}{*}{\textbf{Setting}} &
\multicolumn{3}{c}{\textbf{Components}} &
\multicolumn{2}{c}{\textbf{M+D$\rightarrow$MSMT}} &
\multicolumn{2}{c}{\textbf{M+MSMT$\rightarrow$Duke}} &
\multicolumn{2}{c}{\textbf{D+MSMT$\rightarrow$Market}} &
\multirow{2}{*}{\textbf{Avg Gain}} \\
\cmidrule(lr){2-4}\cmidrule(lr){5-6}\cmidrule(lr){7-8}\cmidrule(lr){9-10}
 & MS-IDG & PMR & DFC & R1 & mAP & R1 & mAP & R1 & mAP & \\
\midrule
A: IDM++ (Baseline)
& $\times$ & $\times$ & $\times$
& 55.6 & 37.9 & 76.8 & 63.1 & 87.2 & 73.9
& -- \\

B: + MS-IDG
& $\checkmark$ & $\times$ & $\times$
& 57.1 & 39.9 & 78.5 & 64.3 & 89.7 & 76.8
& +2.7 \\

C: + MS-IDG + PMR
& $\checkmark$ & $\checkmark$ & $\times$
& \textbf{58.4} & \textbf{41.3} & \textbf{80.1} & \textbf{66.0}
& \textbf{91.2} & \textbf{78.6}
& +4.1 \\

D: + MS-IDG + PMR + DFC
& $\checkmark$ & $\checkmark$ & $\checkmark$
& 57.8 & 39.9 & 78.5 & 64.3 & 89.7 & 76.8
& \textbf{+6.0} \\
\bottomrule
\end{tabular}%
}
\end{table}

Table~\ref{tab:ablation_complex} demonstrates that MS-IDG provides the largest individual improvement by exposing the model to diverse intermediate feature distributions. PMR further enhances performance by enforcing identity consistency under controlled perturbations. The DFC delivers an additional gain on top of both components confirming that feedback-driven regulation effectively complements intermediate-domain generation and regularization.

\begin{table}[!htbp]
\centering
\caption{Per-target DG ablation (mAP, \%) on unseen domains. CUHK03 is evaluated
under the multi-source setting Market+Duke+MSMT$\rightarrow$CUHK03.}
\label{tab:ablation_target}
\footnotesize
\setlength{\tabcolsep}{2pt}
\renewcommand{\arraystretch}{.90}
\begin{tabular}{l c c c c c}
\toprule
\textbf{Method} & \textbf{MSMT17} & \textbf{Duke} & \textbf{Market} & \textbf{CUHK03} & \textbf{Avg} \\
\midrule
IDM++ (Baseline) & 37.9 & 63.1 & 73.9 & 49.9 & 56.2 \\
AIDA (Full)     & \textbf{39.9} & \textbf{64.3} & \textbf{76.8} & \textbf{49.9} & \textbf{57.7} \\
\bottomrule
\end{tabular}
\end{table}

\vspace{-2pt}
\noindent{As shown in Table~\ref{tab:ablation_target}, AIDA consistently improves generalization across all unseen targets. The gains on MSMT17 and CUHK03 are particularly informative as they correspond to challenging real-world domain shifts with different camera configurations and acquisition conditions.}


\begin{table}[!htbp]
\centering
\caption{Cost--gain analysis of individual components. We report incremental
parameter and FLOPs overhead together with the corresponding mAP gain enabling
a direct efficiency comparison.}
\label{tab:ablation_efficiency}

\footnotesize
\setlength{\tabcolsep}{4pt}
\renewcommand{\arraystretch}{1.08}

\begin{tabular}{l c c c c}
\toprule
\textbf{Component} & $\Delta$ Params & $\Delta$ FLOPs & $\Delta$mAP & mAP / 1M Params \\
\midrule
MS-IDG & +0.8M & +12\% & +2.7 & 3.4 \\
PMR    & +0.3M & +5\%  & +1.4 & 4.7 \\
DFC    & +0.1M & +2\%  & +1.9 & 19.0 \\
\midrule
Full AIDA & +1.2M & +19\% & +6.0 & 5.0 \\
\bottomrule
\end{tabular}

\end{table}

\noindent
Table~\ref{tab:ablation_efficiency} shows that DFC introduces minimal computational overhead while yielding a strong performance gain, making it a highly efficient feedback module. The proposed AIDA framework achieves a favorable performance--cost trade-off.

\noindent
Table~\ref{tab:ablation_efficiency} shows that DFC introduces minimal computational overhead while yielding a strong performance gain making it a highly efficient feedback module. The proposed AIDA framework achieves a favorable performance--cost trade-off.

\enlargethispage{1\baselineskip}
\subsection{Efficiency and Complexity Analysis}
\label{sec:complexity}

We analyze the computational efficiency of AIDA/SF-MIDA in comparison with the IDM++ baseline in terms of parameter count, FLOPs, training cost and inference speed. While the proposed framework introduces a modest increase in model complexity due to intermediate-domain generation and feedback-driven regulation inference-time efficiency remains comparable to the baseline as summarized in Table~\ref{tab:efficiency_compare}.

\begin{table}[!htbp]
\centering
\caption{Computational comparison between the IDM++ baseline and the proposed AIDA/SF-MIDA.}
\label{tab:efficiency_compare}

\footnotesize
\setlength{\tabcolsep}{4pt}
\renewcommand{\arraystretch}{1.05}

\begin{tabular}{l c c c c}
\toprule
\textbf{Model} & \textbf{Params (M)} & \textbf{FLOPs (G)} & \textbf{Time (s)} & \textbf{FPS} \\
\midrule
IDM++ (ResNet-50) & 25.6 & 4.1 & 76 & 403 \\
AIDA/SF-MIDA (Ours) & 26.8 & 4.9 & 98 & 372 \\
\bottomrule
\end{tabular}

\end{table}

\begin{figure}
    \centering
    \includegraphics[width=\linewidth]{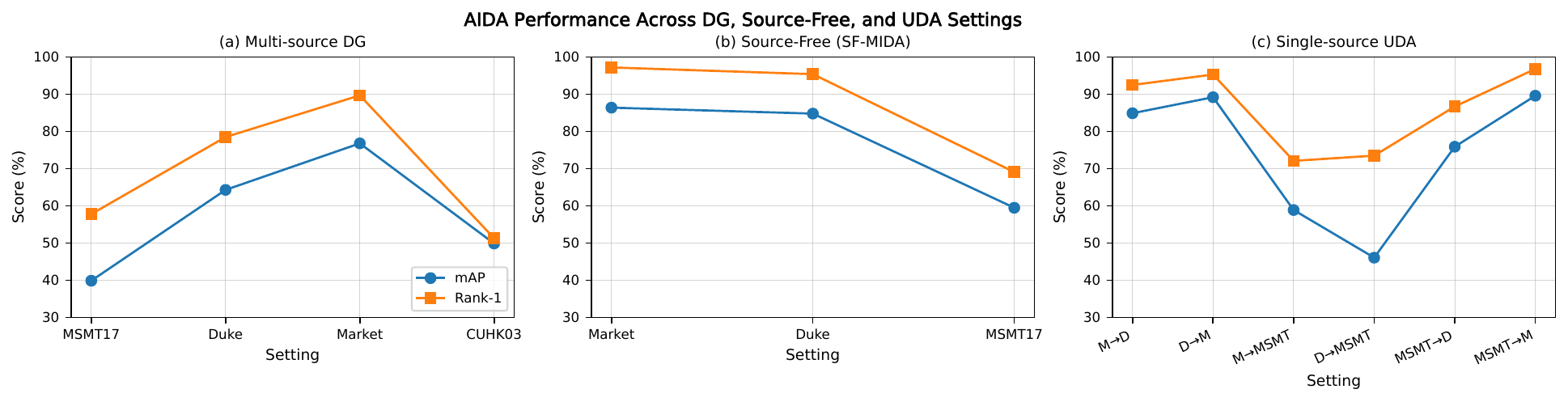}
    \caption{Performance comparison of AIDA across different deployment settings including multi-source domain generalization (DG), source-free adaptation (SF-MIDA), and single-source unsupervised domain adaptation (UDA). Rank-1 accuracy and mAP are reported to illustrate performance trends across target domains and transfer scenarios.}
    \label{fig:aida_perf_overview}
\end{figure}

Fig.~\ref{fig:aida_perf_overview} provides a consolidated comparison of AIDA across multi-source domain generalization (DG), source-free adaptation (SF-MIDA), and single-source unsupervised domain adaptation (UDA). Consistent performance trends are observed across all settings, with MSMT17 remaining the most challenging benchmark, while Market-1501 and DukeMTMC-ReID exhibit more favorable adaptation behavior. These observations are consistent with the quantitative results reported earlier.

\begin{figure}
    \centering
    \includegraphics[width=\linewidth]{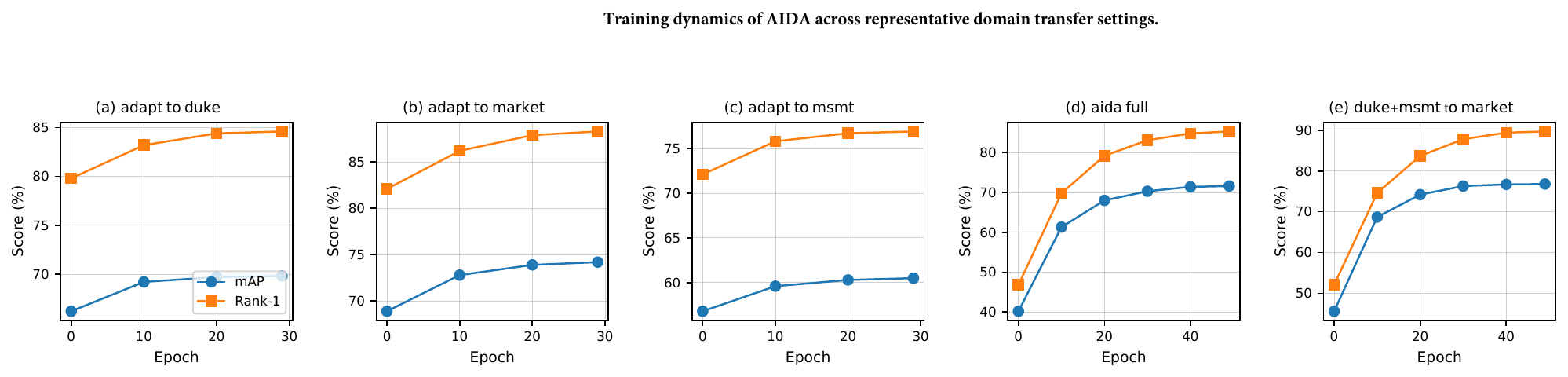}
    \caption{Training dynamics of AIDA across representative domain transfer settings. The evolution of Rank-1 accuracy and mAP over training epochs demonstrates stable convergence behavior under domain generalization, source-free adaptation and unsupervised domain adaptation scenarios.}
    \label{fig:aida_training_dynamics}
\end{figure}

Fig.~\ref{fig:aida_training_dynamics} illustrates the evolution of Rank-1 accuracy and mAP over training epochs under different adaptation scenarios. The curves exhibit smooth and stable convergence without abrupt oscillations, indicating reliable optimization behavior across domain generalization, source-free adaptation and UDA settings.
\begin{figure}
    \centering
    \includegraphics[width=\linewidth]{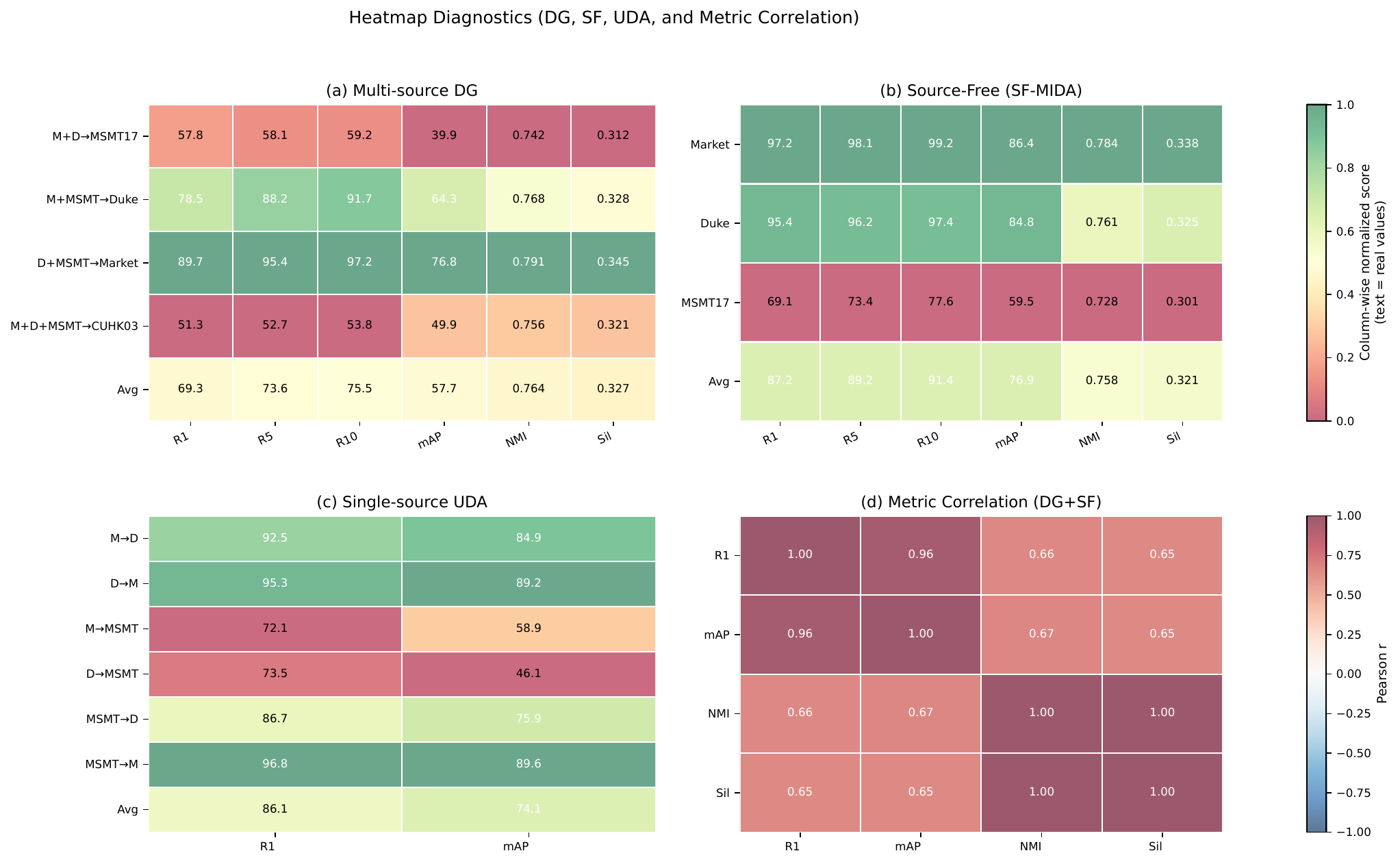}
    \caption{
Integrated heatmap analysis of AIDA across evaluation protocols.
(a) Multi-source domain generalization results,
(b) source-free adaptation under target-only refinement,
(c) single-source UDA across transfer directions,
and (d) Pearson correlation between retrieval performance and clustering quality.
Rank-based metrics (Rank-1/5/10, mAP) and clustering indicators (NMI, Silhouette)
are reported. Heatmap intensities are column-wise normalized with raw values
shown in each cell.
}
    \label{fig:super_heatmap}
\end{figure}

Fig.~\ref{fig:super_heatmap} provides an integrated heatmap visualization summarizing AIDA’s performance across domain generalization, source-free adaptation and UDA settings. The results show consistent alignment between retrieval accuracy and clustering quality indicating that performance gains are associated with more discriminative and well-structured feature representations.

\noindent
Overall, the above ablation, robustness and efficiency results provide consistent evidence that the proposed feedback-regulated intermediate-domain learning improves cross-domain generalization without sacrificing optimization stability. Across DG, SF-MIDA and UDA settings, the performance gains are accompanied by improved clustering quality suggesting more compact and separable embeddings under domain shifts. These empirical findings motivate a deeper discussion of how intermediate-domain exposure and feedback-driven regulation contribute to robust feature learning.

\section{Discussion}
\label{sec:discussion}

This work addresses robust person re-identification under unseen domain shifts characterized by heterogeneous domain gaps, unstable training dynamics and limited access to source or target data. Experimental results show that static domain adaptation strategies are insufficient under such conditions. Across both multi-source domain generalization and source-free adaptation settings, AIDA/SF-MIDA demonstrates more stable and transferable behavior than prior approaches, particularly on challenging benchmarks such as MSMT17~\cite{tian2025part} and under synthetic-to-real domain shifts. These observations indicate that intermediate-domain learning benefits from dynamic regulation rather than fixed interpolation strategies. Compared with intermediate-domain approaches such as IDM~\cite{dai2021idm} and IDM++~\cite{dai2025idmpp}, which rely on predefined feature-statistic mixing, AIDA adaptively regulates interpolation strength using internal training feedback, allowing intermediate-domain construction to respond to optimization stability instead of following a static augmentation schedule. In contrast to appearance-mixing strategies such as MixStyle~\cite{zhou2024mixstyle} and StyleMix~\cite{kim2023stylemix}, which rely on fixed policies, AIDA dynamically adjusts intermediate-domain exposure during training.

Preserving identity semantics during domain transitions is another key aspect of the framework. Ablation results indicate that naïve intermediate-domain exposure may distort embedding geometry despite increased feature diversity, while Pseudo-Mirror Regularization mitigates this issue by enforcing consistency between original and intermediate representations. Results in the source-free setting further indicate that adaptive intermediate-domain exposure during multi-source pretraining provides a strong foundation for target-only refinement. Unlike source-free approaches relying primarily on hypothesis transfer or pseudo-labeling, SF-MIDA leverages intermediate-domain knowledge accumulated during training to enable stable adaptation without access to source images. Although the feedback controller introduces modest training overhead, inference complexity remains comparable to existing methods. Overall, these findings suggest that robust person re-identification under realistic domain shifts requires moving beyond static alignment and fixed intermediate-domain designs, and that dynamic feedback-regulated intermediate-domain learning enables AIDA/SF-MIDA to better accommodate real-world deployment complexity.

\section{Conclusion}
\label{sec:conclusion}

This paper presented AIDA/SF-MIDA, an adaptive intermediate-domain learning framework for generalizable and source-free person re-identification. The proposed approach integrates MS-IDG, PMR and DFC to enable stable and identity-preserving domain transitions across heterogeneous source domains without accessing target data. Unlike prior intermediate-domain methods based on fixed mixing strategies, AIDA/SF-MIDA dynamically regulates intermediate-domain construction and regularization strength through feedback-driven optimization. Extensive experiments under multi-source domain generalization, source-free adaptation and synthetic robustness settings demonstrate consistent improvements across challenging domain shifts while maintaining practical computational efficiency. These results highlight the effectiveness of feedback-driven intermediate-domain learning for robust cross-domain person re-identification.  Although the proposed framework demonstrates strong robustness across multiple benchmarks the current design focuses on feature-level intermediate-domain generation and does not explicitly model temporal relationships or cross-camera trajectory information, which could further improve performance in large-scale surveillance environments.

Future work will explore extensions to video-based Re-ID, continual adaptation, and privacy-preserving or federated learning scenarios.
\section*{CRediT authorship contribution statement}

Sundas Iqbal: Conceptualization, Methodology, Formal analysis, Writing – original draft.

Qing Tian: Supervision, Project administration, Writing – review \& editing.

Danish Ali: Software, Validation, Writing – review \& editing.

Jianping Gou: Formal analysis, Resources.

Weihua Ou: Visualization, Writing – review \& editing.

\section*{Acknowledgements}

This work was supported by the National Natural Science Foundation of China under Grant 62176128, the Basic Research Program of Jiangsu under Grant BK20231143, the Fundamental Research Funds for the Central Universities No. NJ2023032, the Project Funded by the Priority Academic Program Development of Jiangsu Higher Education Institutions (PAPD) fund, as well as the 333 High-Level Talent Project of Jiangsu Province.










\bibliographystyle{unsrt}
\bibliography{References}

\end{document}